# Multi-modal data fusion of Voice and EMG data for Robotic Control

Tauheed Khan Mohd
*EECS Department*
*The University of Toledo*
Toledo, OH, USA
tauheed.khanmohd@utoledo.edu

Jackson Carvalho
*EECS Department*
*The University of Toledo*
Toledo, OH, USA
jackson.carvalho@utoledo.edu

Ahmad Y Javaid
*EECS Department*
*The University of Toledo*
Toledo, OH, USA
ahmad.javaid@utoledo.edu

*Abstract*— **Wearable electronic equipment is constantly evolving and is increasing the integration of humans with technology. Available in various forms, these flexible and bendable devices sense and can measure the physiological and muscular changes in the human body and may use those signals to machine control. The MYO gesture band, one such device, captures Electromyography data (EMG) using myoelectric signals and translates them to be used as input signals through some pre-defined gestures. Use of this device in a multi-modal environment will not only increase the possible types of work that can be accomplished with the help of such device, but it will also help in improving the accuracy of the tasks performed. This paper addresses the fusion of input modalities such as speech and myoelectric signals captured through a microphone and MYO band, respectively, to control a robotic arm. Experimental results obtained as well as their accuracies for performance analysis are also presented.**

## I. INTRODUCTION

Interaction with machines and robots has been a result of the industrial revolution since the last century. The fourth industrial revolution is comprised of technologies that fuse the physical, digital and biological worlds, and it is impacting all disciplines, economies, and industries [1]. It emphasizes designing robots capable of performing tasks while accepting inputs from different modalities such as speech, gestures, & also the utilization of peripheral devices. The mode of interaction is selected according to the conditions prevailing in the environment. For example, in a noisy environment of a manufacturing plant, a robot may be operated through gestures and peripheral devices excluding speech or by using a combination of available modalities. Robots are an integral part of our daily lives; they have been widely used in the automotive, healthcare, manufacturing, and aerospace industries. The manufacturing industry has been shifting to automate processes, as a way if improving quality, reducing human intervention which is more error prone, and, perhaps the largest advantage, relates to the benefits from the lack of fatigue a robot experiences relative to a human worker.

This paper addresses the challenge faced by users required to control a robot using a set of dedicated controls. In this scenario, we are fusing electromyography data captured from human limbs with human speech. A robot is capable of performing predefined tasks, for example, holding a spare part from a specific location in a 3D environment and inserting it into a machine at a specific location. Previously, this task was performed by robots using a single modality, i.e., controlled through a joystick or a programmable board attached to a robot. Now, we are trying to make the system more accurate by utilizing a fusion of modalities. The design will incorporate voice commands along with captured muscle movements interpreted via EMG data from the arm band.

The rest of this paper is organized as follows: Section 2 discusses related work involving multimodal systems with their applications. Section 3, Methodology, addresses both the hardware and the software used, Section 4 provides the results of data fusion based on the experiments performed. Finally, Section 5 discusses various new avenues of research, as well as emerging trends in data fusion using EMG.

## II. RELATED WORK

New technology often brings with it the idea that machines are not only to be used for the specific predefined purpose, but they can also require the interaction with humans[2]. Human-robot communication is possible through two methods.

- Accepting user input from peripheral devices which are independent of each other, and

- Accepting user input through different modalities and fusing them as a way of obtaining the semantics associated with the actions of the user.

The second approach is the focus of this paper. In 2005, a system was designed which accepts input in the form of speech, keystrokes, and gestures. This system was able to resolve ambiguous inputs and prioritize them [3]. Fusion of multiple inputs is used in several areas of application, and its scope is not only confined to robots, but it also reaches to applications such as authentication systems where fusion could be utilized by say, for instance, combining voice recognition and facial detection. A system was designed in 1999 which was able to authenticate a user by comparing inputs against a pre-populated database [4]. The latest version of Microsoft Windows, Windows 10, is capable of authenticating users through a webcam attached to the computer system [5], though it is a unimodal system that could be enhanced with more modalities to improve its accuracy and make it less vulnerable to outside attacks or spoofing. The use of EMG data in fusion is rarely encountered; one such application was implemented to control electronic musical devices through EMG and relative position sensing [6]. The idea of multimodal data fusion has been implemented in industrial robots using the Microsoft Kinect and sensor hardware called Asus Xtion Monitor by capturing hand movements detected by two Leap Motion sensors and performing the resulting mapped actions on a robotic arm [7]. Human-robot interaction during the last five years has largely been performed using the Microsoft Kinect; very few multimodal system designs have used EMG data to fuse with speech, text, and other modalities.

The Microsft Kinect is capable of capturing both voice and gestures only on a standalone basis. Moreover, there is no ability to capture EMG data of human limbs using the Kinect. This limitation led us to another niche technology called the MYO sensor arm band. We have decided to use it as one of the modalities and perform fusion to enhance both its accuracy and performance. MYO arm band being an open source software provides avenues to customize gesture and use them in devices used in daily life, for example, controlling a wheel chair, rotating a door knob, etc [8]. The results achieved with other experiments to evaluate the accuracy of MYO comes out to 87.8 to 89.38% which provides us avenues for improvement [9]. MYO band is used in an experiment to perform both search and select operations on a computer, and the average score for evaluation is analyzed, the researchers conclude that after adapting the limbs temperature the MYO band constantly performs with a similar number of scores [10]. It is our belief that is this the first research work undertaken that uses EMG data to capture gestures using MYO armband sensors for Multimodal data fusion.

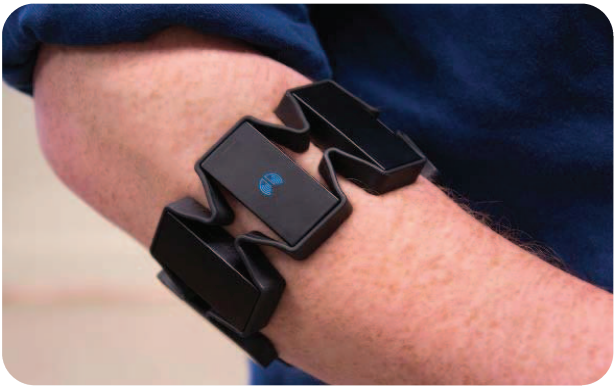

*Figure 1 MYO Arm Band*

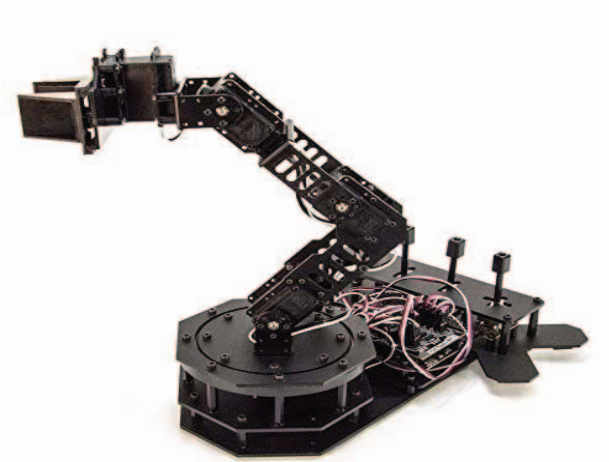

*Figure 2 Arduino based robotic arm*

*Table 1 Mapping of Gestures with Arduino Boards*

| Fist | Pin 3 |
|---|---|
| Wave In | Pin 4 |
| Wave Out | Pin 5 |
| Finger Spread | Pin 9 |
| Double Tap | Pin 10 |

*Table 2 Preliminary results for of Muscle sensor MYO band*

| Gesture | Wrong/Missed gesture % | Correct % |
|---|---|---|
| Wave Out | 9.5 | 90.5 |
| Wave In | 9.1 | 90.9 |
| Fist | 13.6 | 86.4 |
| Double Tap | 20.6 | 79.4 |
| Finger Spread | 14.5 | 85.5 |

## III. METHODOLOGY

The robots first introduced to the market were relatively simple, most of them requiring a teaching phase and programming. More recently, robots become dynamic, sophisticated, and much more capable than before [11]. Along with this sophistication came increasing demands to perform complex tasks which require both accuracy and precision. The standalone robot in the experiment introduced in this paper showed ample room for improvement in both robustness and accuracy. Hence it was decided to improve the accuracy of a robotic arm by the use of multi-modal data fusion. The experiment emphasizes the conversion of input data through different channels into a single format which is understood by the robotic arm through mediation. The input modalities used are speech and gesture.

### A. Hardware

The system is designed for the experiment described in this paper is composed of the following components: An Arduino based robotic arm, and an MYO arm band. These devices are illustrated in Figure 1 and Figure 2 respectively. The robotic arm used is manufactured by Trossen Robotics. The robot used in the experiment described here is called 'RobotGeek Snapper Arduino Robotic Arm,' and it contains five servo motors. The robotic arm is controlled by an electromyography data-based arm band called an MYO. Figure 1 depicts the usage of the MYO band from which data is captured and manipulated to perform actions on various Arduino-based devices. The arm band is capable of capturing five gestures: Fist, Wave Left, Wave Right, Double Tap, and Fingers Spread as shown in Figure 3. This arm band provides the ability to customize an open library and perform actions per users need. The robotic arm

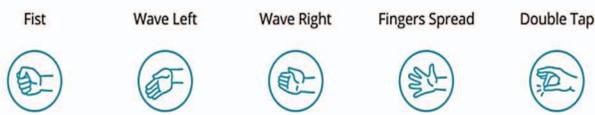

Fist    Wave Left    Wave Right    Fingers Spread    Double Tap

*Figure 3 Gestures available with MYO band*

used is comprised of Arduino Duemilanove and Diecimila boards for accepting input through USB. The Arduino board is connected to the robot with pins defined for each specific motor as shown in Figure 2. A high precision wireless H800 headset from Logitech is used for capturing the speech input.

### B. Software

The software is implemented using C# and C++ programming languages. C++ is used to implement the MYO API while C# is used for providing speech input using Microsoft's speech engine. Initially, the robotic arm was operated using gestures only. The MYO arm band was used to support the communication with the robot. It was concluded that the precision of the armband was not very high. To improve the accuracy of the commands communicated fusion of the input data was introduced. The speech modality was then combined with the set of gestures previously defined. The MYO band provides an API to control the Arduino board which is connected to the robotic arm as shown in Figure 2. The Arduino board consists of 14 pins and provides the capability to connect each servo motor of the robotic arm.

### C. Experimental Setup

The experimental system is designed using a client-server paradigm. The MYO arm band consists of eight sensors which capture the muscle movement. Such sensors compare and match hand movements to gestures defined in MYO. For example, the Wave Out gesture listed in Table 2 is performed by moving the hand in a vertical orientation toward the right as shown in the third gesture of Figure 3. The MYO program is functioning as the server while the Microsoft speech program is designed as the client. As mentioned above, the accuracy of capturing gestures is not very high, which tends to result in matching the hand movement to the wrong gesture, e.g., Wave Out may be captured as Wave In. Moreover, the band sometimes fails to capture a gesture entirely. Both of these cases are considered errors. The MYO API allows it to be customized according to the project needs and the prototype has thus implemented threads responsible for listening to gestures.

If a gesture is missed by the band, voice commands compensate for the missed input through human speech. Processing of speech input is implemented in a client component which sends commands to the server (MYO API). Priority is given to the MYO band, but in the case of an error, speech recognition activates and helps in improving the accuracy of controlling the robot. Fusion is thus performed in the order of priority. Gestures are given the highest priority. In case of a failure to capture the input, voice commands are used to compensate and serve as the only input. Priority-based fusion is used in other domains including medical systems and tends to improve its accuracy significantly [12].

The speech is fused with EMG input received from MYO, which enables the robot to work precisely as per the user's input command. When the user performs gestures using his/her arm, the input message is transmitted from the MYO band to the Arduino, and as a result, it moves the specific servo motor. The fused input sets the corresponding Arduino pin to high, i.e., 1, which then moves the robot. In the prototype constructed, five different Arduino pins were linked to various gestures as shown in Table 1. Through the combination of gestures and speech, users should be able to control the robotic arm precisely and accurately. The results are shown in the next section.

### IV. RESULTS AND DISCUSSION

A performance evaluation was executed to quantify how accurate the modalities are individually, and thus we tested them separately. The Microsoft Speech API was tested using the simple speech commands such, say, for example, "move right," "move left" etc. The complete list of speech commands is illustrated in Table 3. The experimental results show the scope of improvement as the error for the speech API lies between 8.9 to 34.2%. Similarly, the MYO band results were captured to quantify accuracy and to find the scope of improvement. The preliminary results have shown the MYO band has a scope of improvement. The error rate lies between 9.1 to 20.6%. The data has been collected by experimenting 10 times, with each experiment having 100 gestures performed and then calculating the average percentages shown in Table 2. An error for the experiment occurs when a gesture is either missed or captured wrong. The trials have been performed in laboratory conditions. The MYO arm band is capable of adapting to specific human limbs and improves its output once it has been trained completely. The arm band sensors become warm up shortly after being put on, thus adapting to body temperature and accurately recognizing gestures after 1-2 minutes. If one's arm is cold, the sensors are unable to capture gestures accurately. The Microsoft API results are evaluated by speaking a command, and if the command is captured incorrectly, the instance is marked as an error. Incorrect capture is defined as the resulting string being captured twice or having extraneous words or characters added to it. Multimodal data fusion of voice and gesture using the MYO band improves the system performance significantly. The experimental results are shown in Table 4. After implementing fusion in the robotic arm, the error rate is reduced to 5.2% which is an average of all errors. The variance of error % is shown in Figure 4. The errors are mostly due to reading the wrong gesture, e.g., finger spread is sometimes captured as fist which leads to an error. Experiments are performed on all the five fusion input testing 200 times each, and the percentage is calculated respectively.

*Table 3 Preliminary results for Microsoft Speech used by non-native speaker.*

| Command | Microsoft Speech API (Wrong output) | Correct |
|---|---|---|
| Move Right | 10 | 90 |
| Move Left | 34.2 | 65.8 |
| Move Up | 8.9 | 91.1 |
| Move Down | 22.5 | 77.5 |
| Move Gripper | 14.1 | 85.9 |

*Table 4 Fusion results with Error % & Variance*

| Fusion Operation | 50 | 100 | 150 | 200 | Error % | Variance |
|---|---|---|---|---|---|---|
| Move Gripper & Double Tap | 7 | 2 | 2 | 4 | 7.5 | 5.58 |
| Move Down & Fist | 3 | 1 | 2 | 2 | 4 | 0.67 |
| Move Up & Finger spread | 3 | 3 | 2 | 2 | 5 | 0.33 |
| Move Left & Wave left | 0 | 3 | 2 | 2 | 3.5 | 1.58 |
| Move Right & Wave out | 3 | 3 | 4 | 2 | 6 | 0.67 |

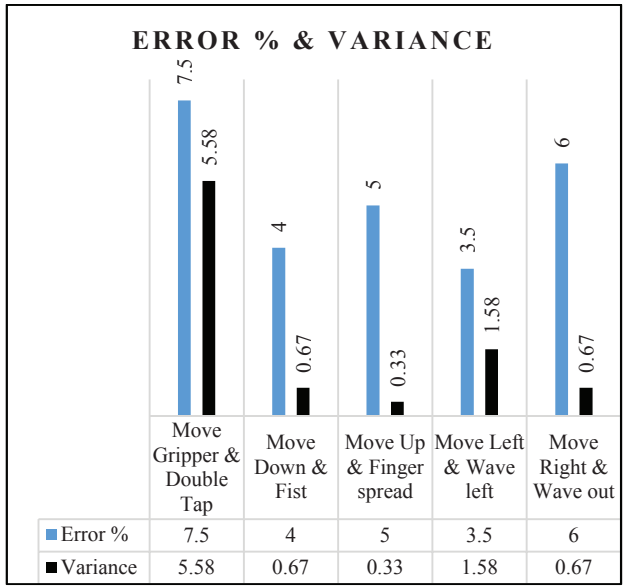

| | Move Gripper & Double Tap | Move Down & Fist | Move Up & Finger spread | Move Left & Wave left | Move Right & Wave out |
|---|---|---|---|---|---|
| ▇ Error % | 7.5 | 4 | 5 | 3.5 | 6 |
| ▇ Variance | 5.58 | 0.67 | 0.33 | 1.58 | 0.67 |

*Figure 4 Error % deviation of fused Inputs*

## V. LIMITATIONS

Speech and electromyography data were the modalities used in the system constructed. The MYO band used for capturing the EMG data is capable of recognizing five gestures. This limited the number of operations that could perform on the robotic arm. The second challenge lay in capturing the speech commands using the Microsoft Speech API. Non-native speakers of the English language will face difficulties and challenges to approximate his/her accent to that of a native speaker's. This created difficulty in conveying commands correctly. There are four areas which need to be worked on and improved regarding human-robot interaction. They are speech localization, language understanding, dialogue management, and speech synthesis [13]. Also, as the ultimate goal of this research is to improve accuracy, the approach here described maps commands to all possible options that the Microsoft Speech API recognizes as valid (for example "move right" sometimes gets recognized as "override" – an incorrect response). This provided us with a way to quantify the accuracy of the system. We prepared a many-to-one mapping of all these possible combinations to a particular voice command.

## VI. CONCLUSIONS

The experimental results displayed above prove that by the inclusion of speech input modality the accuracy of the MYO band can be improved significantly. While using the modalities separately, the accuracy was 86.54% and 82.06% for the MYO band and Microsoft Speech API, respectively. After fusion of the inputs, accuracy improved to more than 95.92%. Our future work includes the development of a prototype in which the system can perform a fusion of more than two input modalities and perform tasks after interpreting the semantics of the input provided. Thus far there is no system which takes input from the user in the form of speech, text, and gesture and executes a task on robotic arm using the MYO band dynamically. The next planned implementation will add components able of capturing brain signals. The system should be able to fuse the modalities and select a meaningful operation that is then performed on a device.